\documentclass[journal]{IEEEtran}
\let\emph\textit
\usepackage{cite}
\usepackage{amssymb,amsfonts}
\usepackage{algorithmic}
\usepackage{amsmath} 
\usepackage{autobreak}
\usepackage{graphicx}
\usepackage{textcomp}
\usepackage[numbers]{natbib}
\usepackage{algorithm}
\usepackage{enumerate}
\usepackage{subfigure}
\usepackage{multirow}
\usepackage{autobreak}
\usepackage{booktabs}
\usepackage{color, soul}
\usepackage{url}

\usepackage[T1]{fontenc}

\ifCLASSINFOpdf
\else
\fi
\usepackage{amsmath}
\usepackage[cmintegrals]{newtxmath}
\begin{document}

%
\title{Optimizing Deep Neural Networks through Neuroevolution with Stochastic Gradient Descent}

%
%
%
\author{Haichao Zhang,
Kuangrong Hao,~\IEEEmembership{Member,~IEEE,}
Lei Gao,~\IEEEmembership{Member,~IEEE,}
Bing Wei,~\IEEEmembership{Student Member,~IEEE,}
Xuesong Tang
\thanks{This work was supported in part by the National Natural Science Foundation of China (nos. 61806051, 61903078), and Natural Science Foundation of Shanghai (20ZR1400400, 19ZR1402300).}
\thanks{Corresponding author: Kuangrong Hao, Lei Gao}
\thanks{H.-C. Zhang, K. Hao ,B. Wei and X.-S. Tang are now with the College of Information Science and Technology, Engineering Research Center of Digitized Textile and Apparel Technology, Ministry of Education, Donghua University, Shanghai 201620, P. R. China (e-mail: krhao@dhu.edu.cn). L. Gao is now with CSIRO, Waite Campus, Urrbrae, SA 5064, Australia.}
}

\maketitle

\begin{abstract}
  Deep neural networks (DNNs) have achieved remarkable success in computer vision; however, training DNNs for satisfactory performance remains challenging and suffers from sensitivity to empirical selections of an optimization algorithm for training. Stochastic gradient descent (SGD) is dominant in training a DNN by adjusting neural network weights to minimize the DNN’s loss function. As an alternative approach, neuroevolution is more in line with an evolutionary process and provides some key capabilities that are often unavailable in SGD, such as the heuristic black-box search strategy based on individual collaboration in neuroevolution. This paper proposes a novel approach that combines the merits of both neuroevolution and SGD, enabling evolutionary search, parallel exploration, and an effective probe for optimal DNNs. A hierarchical cluster-based suppression algorithm is also developed to overcome similar weight updates among individuals for improving population diversity. We implement the proposed approach in four representative DNNs based on four publicly-available datasets. Experiment results demonstrate that the four DNNs optimized by the proposed approach all outperform corresponding ones optimized by only SGD on all datasets. The performance of DNNs optimized by the proposed approach also outperforms state-of-the-art deep networks. This work also presents a meaningful attempt for pursuing artificial general intelligence.
\end{abstract}

\begin{IEEEkeywords}
  Deep neural networks ,  stochastic gradient descent ,  neuroevolution , image classification
\end{IEEEkeywords}

%
\IEEEpeerreviewmaketitle

\section{Introduction}

In recent years, deep neural networks (DNNs) have been making ground-breaking advances in the field of computer vision, such as image classification \cite{R1,R2,R3}, object detection \cite{R4,R5}, and image segmentation \cite{R6,R7}. Backpropagation is widely used in optimizing DNNs for supervising learning \cite{R9}. In training a DNN, backpropagation calculates the gradient of the loss function and minimizes the loss by adjusting the weights of the network. There are a number of gradient-based optimization strategies to accomplish backpropagation tasks and stochastic gradient descent (SGD) is a typical and dominant one \cite{R10}. However, even with SGD, the loss function of a DNN is still difficult to converge to a satisfying minimum and this strongly affects the DNN performance. Consequently, some efforts have been made in developing variants of SGD to seek better training/optimization performance. Dauphin et al. \cite{R11} proposed RMSProp to improve SGD, which solves the optimization of the loss function in the presence of saddle points through an adaptive learning rate scheme based on an equilibration preconditioner. Similar to RMSProp, Diederik et al. \cite{R12} proposed an optimizer based on adaptive estimates of lower-order moments to assist SGD in adaptively adjusting the learning rate and improving the convergence. Although these efforts improved the SGD and achieve good training performance to some extent, how to obtain well-trained DNNs remains challenging.

On the other hand, taking inspiration from biological mechanisms and designing novel frameworks/algorithms is an effective approach for problem-solving in the field of DNNs \cite{R13,R1}. Inspired by biologically evolutionary processes that produced natural brains, neuroevolution is an alternative approach to train neural networks with evolutionary algorithms \cite{R15}. Neuroevolution retains a population of potential solutions in the process of evolutionary search, supporting large-scale exploration and parallelization. It also offers marked capabilities that gradient-based methods lack, such as learning building blocks \cite{R16} and hyperparameters \cite{R17} of DNNs. In the early research, neuroevolution mainly optimized the network topology by simply varying the weights in the connection matrix \cite{R18}. The focuses of some studies then turned to explore the coding of more complex network topologies \cite{R19}. To make evolving increasingly complex topological structure more effective, Stanley and Miikkulainen \cite{R20} solved the problem of crossing over variable topologies through the historical marking of neural networks, which used a mechanism called speciation to prevent the premature disappearance of newly generated structures. With the enhancement of modern computing resources and the development of DNNs, neuroevolution has been applied in optimizing DNNs and achieved creditable optimization performance. These efforts focused on optimizing network weights \cite{R21}, architecture \cite{R22,R23}, and hyperparameters \cite{R24,R25}. Lehman et al. \cite{R26} developed an approach to introduce safety mutations into the output gradient. This approach successfully evolved network weights of more than 100 layers and improved the performance of neuroevolution. Another method of evolving neural networks was proposed to run gradient-based reinforcement learning as the engine of crossover and mutation in the evolution process \cite{R27}. There were also some research efforts \cite{R28} showing the combination of evolutionary algorithms with policy gradients and QLearning, creating random perturbations to drive the iteration of the algorithms, and then optimizing the weights of DNNs through backpropagation.

Modern neuroevolution offers key capabilities that are normally not available to SGD (such as large-scale and parallel exploration of optimal DNNs, power of encoding, meta-learning, and architecture search), while SGD has been proven remarkably effective for optimizing DNNs. However, there are few studies that combined the merits of the two approaches for training neural networks. Recently, Cui et al. \cite{R29} designed an evolutionary stochastic gradient descent (ESGD) to optimize DNNs by alternating between the SGD step and the evolution step. A coevolution mechanism was applied in the ESGD where candidate individuals with different optimizers were regarded as competing species. ESGD is a novel and effective attempt to optimize DNNs by combining SGD with neuroevolution. However, some primary limitations remain, for example, the loss function of a DNN optimization algorithm based on ESGD uses only partial principled merit of neuroevolution, which inadvertently result in the difficulty for the loss function to converge to a satisfactory minimum.

Therefore, through the heuristic black-box search strategy based on individual collaboration in neuroevolution, this paper proposes a novel approach that combines neuroevolution with stochastic gradient descent (NE-SGD) for optimizing DNNs. Specifically, we first converge the loss function of a DNN to a minimum using SGD in the NE-SGD framework. By encoding the weights of the DNN, candidate solutions are generated. Each individual solution is considered a "species" of the population \cite{R30,R31} in the process of neuroevolution. Then, the encoded individual is converged to a minimum through SGD once more and the test accuracy is obtained, which is used as the fitness value of the individual in neuroevolution. In this way, after the initial population is created, the optimization iterations begin and the population continues to experience crossover, mutation, evaluation, and natural selection. During any generation of evolution, individuals are optimized independently in the SGD step and interact with each other in the evolutionary step. The above combination of SGD and neuroevolution largely minimizes the loss function. We also develop a hierarchical cluster-based suppression algorithm to improve the  diversity of the population.

This work adopts convolutional neural networks, one of the most popular DNN architectures, to demonstrate the effectiveness of our approach in optimizing DNNs. The main contributions of this paper are presented below:

\par

\begin{enumerate}[1)]
    \item Combining merits of both the gradient-free neuroevolution and SGD, a called EN-SGD approach is proposed for optimizing DNNs. The approach enables evolutionary search, parallel exploration, and an effective probe for optimal DNNs. It is also a meaningful attempt for pursuing artificial general intelligence.
    \item A suppression method based on hierarchical clustering is proposed to improve the over updates of similar weights among individual solutions in EN-SGD.
    \item The effectiveness of the proposed approach is validated on four datasets: CIFAR10, CIFAR100 \cite{R32}, SVHN \cite{R33}, and Aliyun10500 \cite{R44}. The experiments on four DNN models (ResNet18, ResNet34 \cite{R35}, DenseNet121 \cite{R36}, and DPN92 \cite{R37}) show the effectiveness of the proposed NE-SGD approach.
  \end{enumerate}

\par

The rest of this paper is organized as follows: the second section presents the detailed implementation of the proposed EN-SGD approach. In order to validate the effectiveness and efficiency of the
proposed approach, the experiment designs are presented in the third section. The fourth section provides the experimental results of four public-available datasets on some models. Our research efforts are concluded in the fifth section.

\section{Methods}

The combined NE-SGD framework is first presented in this section for optimizing DNNs, then, key details such as how a DNN is encoded, evolutionary operations, and a hierarchical clustering-based suppression technique for increasing population diversity are given.

\subsection{A combined Neuroevolution-SGD framework for optimizing DNNs}

\begin{figure*}
\centering
\includegraphics[width=18cm]{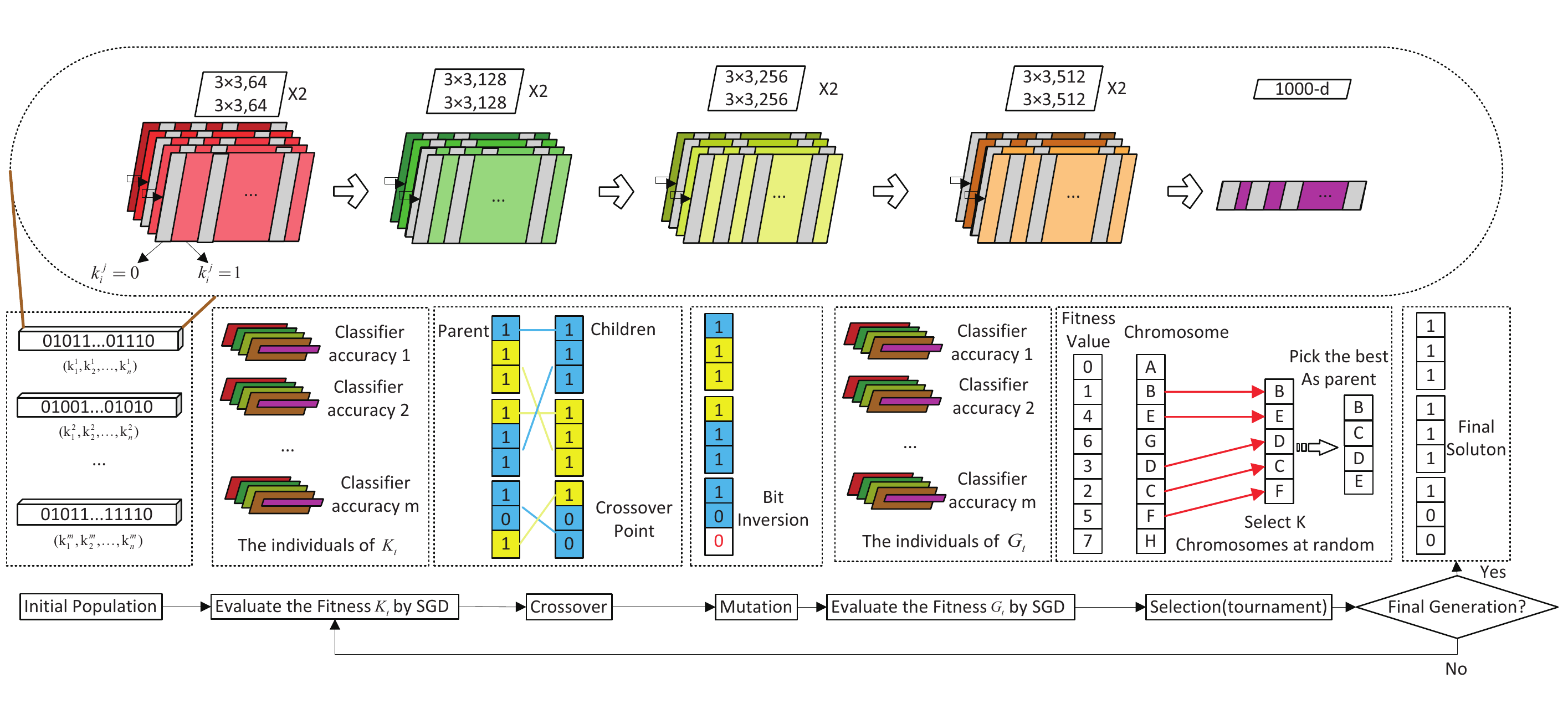}
\caption{The proposed NE-SGD framework for optimizing a basic ResNet18 network. The rounded rectangle shows the decoding implementation of the ResNet18 network. The solid colored region maps $k_{i}^{j}=1$ in the binary code. The gray area maps $k_{i}^{j}=0$ in the binary code. Each ResNet18 network is represented an individual in the evolutionary algorithm. The test accuracy of the model is used as the fitness value of an individual. The parental individual is selected using the tournament method for further crossover and mutation operations. The above process is iterated until the stop criteria are met.}
\label{fig:picture001}
\end{figure*}

The proposed NE-SGD framework is composed of four key steps. First, the loss function of a DNN is optimized by SGD. Next, based on optimized weights by SGD, individual solutions are generated (each individual represents a set of partial network weights for updating) according to the coding approach in neuroevolution. Third, a set of evolutionary operations (such as crossover and mutation) are applied to the population. The fitness values of all individuals are first optimized by SGD and then evaluated and used for natural selection. A hierarchical clustering method is used to improve population diversity. Finally, the best individual is obtained as the final optimized network model. Fig. \ref{fig:picture001} demonstrates the encoding process and the above four steps of the proposed NE-SGD framework using ResNet18. As shown in Fig. \ref{fig:picture001}, an initial population of a DNN (here we use a convolutional neural network—the ResNet18 network as an example) that contains binary-coded individuals (the coding approach is presented in Subsection \uppercase\expandafter{\romannumeral2}-B) is created and optimized using SGD. All individuals experience evolutionary operations, such as crossover and mutation (see Algorithm \ref{algorithm1}). In the neuroevolution step, the ResNet18 network represented by each individual is further optimized by SGD, and the test accuracy is used as the fitness value (see Subsection \uppercase\expandafter{\romannumeral2}-C for details). During the selection process in neuroevolution, the proposed hierarchical cluster-based suppression algorithm is used to avoid the overly similar weight updates among individuals in the population and improve the diversity of the population (see Algorithm \ref{algorithm2} and Subsection \uppercase\expandafter{\romannumeral2}-D for more details).

\subsection{Network Encoding Strategy}
We use a convolutional neural network (CNN) as an example to demonstrate how a DNN is encoded. In this work, a binary encoding method is used to encode network parameters under a constraint. The network structure parameters can be represented in blocks. In each block, the size of the weight parameters keeps the same, and the connection state between these parameters remains unaltered. In the backpropagation process of NE-SGD, the weights are subject to local optimization under various conditions such as random initialization. Considering the above situation, the weight of the entire network will be encoded according to the block of the neural network.

Therefore, a network parameter can be encoded into a fixed-length binary string. The network consists of $n$ blocks, and the width of the coded space is $n$. The parameters of the neural network are defined as $N =\{x_1,x_2,...,x_n\}$. The binary code is represented as the initialization of the trainable parameter block in the network. Thus, the population can be defined as $K=\{(k_{1}^{1},k_{2}^{1},...,k_{n}^{1}),(k_{1}^{2},k_{2}^{2},...,k_{n}^{2}),...,(k_{1}^{m},k_{2}^{m},...,k_{n}^{m})\}$, where $m$ represents the number of individuals $k_{i}^{j}\in \{0,1\},i\in \{1,2,\ldots ,n\},j\in \{1,2,\ldots ,m\}$ . The parameters at the minimum $\alpha$ are defined as $P^{\alpha}$, and $P^{\alpha} = \{P^{\alpha}_1,P^{\alpha}_2\}$, $P^{\alpha}_1 = \{x_i|k_i =1\}$, $P^{\alpha}_2 = \{x_i|k_i =0\}$. The reinitialized parameters in the neural network at the minimum $\alpha+1$  are defined as $P^{\alpha+1}_3 = \{x_i|k_i =0\}$ and $P^{\alpha} = \{P^{\alpha}_1,P^{\alpha}_2\}$. The function $f(x)$ is defined to map $P^{\alpha}_2$ to $P^{\alpha+1}_3$. The formula for the trainable parameter block $x_n$ of encoding is presented in Equation (\ref{equation1}).

\begin{equation} \label{equation1} 
f\left( {{x_n}} \right) = \left\{ \begin{array}{l}
{\rm{initialize}}{\kern 1pt} {\kern 1pt} {\kern 1pt} {\kern 1pt} {\kern 1pt} {\kern 1pt} {\kern 1pt} {\kern 1pt} {\rm{if}}{\kern 1pt} {\kern 1pt} {\kern 1pt} {\kern 1pt} {k_i^j} = 0\\
{\rm{retain}}{\kern 1pt} {\kern 1pt} {\kern 1pt} {\kern 1pt} {\kern 1pt} {\kern 1pt} {\kern 1pt} {\kern 1pt} {\kern 1pt} {\kern 1pt} {\kern 1pt} {\kern 1pt} {\kern 1pt} {\kern 1pt} {\kern 1pt} {\kern 1pt} {\kern 1pt} {\kern 1pt} {\kern 1pt} {\kern 1pt} {\kern 1pt} {\rm{if}}{\kern 1pt} {\kern 1pt} {\kern 1pt} {\kern 1pt} {k_i^j} = 1
\end{array} \right.\end{equation}

Fig. \ref{fig:picture001} illustrates the coding method in EN-SGD on ResNet18 \cite{R35}. The big rounded rectangle represents the network structure diagram formed by binary code decoding. The gray and solid-colored blocks in the rounded rectangle represent the initialization of parameters and reserved parameters, respectively. Each individual decodes a network according to a binary decoding rule, and the test accuracy value of the neural network is used as the fitness value.

\subsection{Evolutionary Operations}

The evolutionary process in EN-SGD can be expressed in Algorithm \ref{algorithm1}, which is consisted of three parts. First, the population is randomly initialized with the size of (see line 1), and all individuals in the population are evaluated for fitness (line 2). Then, all individuals participate in the process of neuroevolution with the maximum iteration number $T$ (lines 2-13). Finally, the best binary code is produced from the final population. During the evolutionary process, a new offspring is generated from selected parents with the crossover and mutation operations, while the parents are selected by the binary tournament selection (lines 6-8). After the fitness of the resulting offspring is assessed, a new population is selected from the current population containing the current individuals and the generated offspring  (lines 10-11). The new population serves as the parent solution for survival to the next evolutionary process. NE-SGD follows the standard process of genetic algorithms as above details but differs in the method of calculating fitness. Next, we will introduce the process of calculating fitness in detail.

\begin{algorithm}[h]
\caption{Evolutionary operations and process in NE-SGD}
\label{algorithm1}
\begin{algorithmic}[1]
\REQUIRE The population size $m$, the maximal generation number $T$, the crossover probability $\mu$, the mutation probability $\nu$, and the reference dataset $\mathcal{D}$.

\STATE Randomly initialize a population $K = \{ (k_1^1,k_2^1,...,k_n^1),(k_1^2,k_2^2,...,k_n^2),...,(k_1^m,k_2^m,...,k_n^m) \}$ with the size of $m$, and the size of the parameter block in the neural network is represented by $n$;

\FOR{$t = 1,2,3,...,T$}
\STATE Evaluate the fitness of individuals in $K_t$ by SGD;
\STATE ${{G}_{t}}\leftarrow \phi $;
    \WHILE{$G_t < n $}
        \STATE $g_1,g_2 \leftarrow $ Select two parent individuals from $K_t$ by use binary tournament selection;
        \STATE $k_1,k_2 \leftarrow$ Generate two off springs by applying the crossover operation to $g_1$ and $g_2$ with the probability $\mu$ and then applying a mutation operation with the probability $\nu$;
        \STATE $G_t \leftarrow G_t \cup k_1 \cup k_2  $;
    \ENDWHILE
\STATE Evaluate the fitness of individuals in $G_t$;
\STATE $K_{t+1} \leftarrow $Select $n$ individuals from $K_t \cup G_t$ by environmental selection;
\STATE $t \leftarrow t+1 $;
\ENDFOR
\ENSURE The best binary code of model.
\end{algorithmic}
\end{algorithm}

Individual fitness provides a quantitative measure of how well individuals are adapted to the environment and it is calculated based on the information encoded by the individuals and the task at hand. First of all, the loss function of a DNN is converged to a minimum $\alpha$ by SGD, and the weights of a DNN are represented as:

\begin{equation}
    P^{\alpha} = \{P^{\alpha}_1, P^{\alpha}_2\}
    \label{equation2}
\end{equation}

Based on Equation (\ref{equation1}) and the individuals $(k_1^j,k_2^j,...,k_n^j)$, the weights  $P^{\alpha} = \{P^{\alpha}_1, P^{\alpha}_2\}$ are mapped to $P^{\alpha+1} = \{P^{\alpha}_1,P^{\alpha+1}_3\}$: 

\begin{equation}
    \{P_{1}^{\alpha },P_{3}^{\alpha +1}\}=f(P_{1}^{\alpha },P_{2}^{\alpha })
    \label{equation3}
\end{equation}

where $P^{\alpha+1} = \{P^{\alpha}_1,P^{\alpha+1}_3\}$ represents the weights of the DNN updated by Equation \ref{equation1} . Then, the network with the weights $P^{\alpha+1} = \{P^{\alpha}_1,P^{\alpha+1}_3\}$ is converged to the minimum $\beta$ on training data. Thus, the network is trained with the weights $P^{\alpha+1} = \{P^{\alpha}_1,P^{\alpha+1}_3\}$:

\begin{equation}
    P^{\beta} = \{P^{\beta}_1,P^{\beta}_3\}
    \label{equation4}
\end{equation}

where $g(\cdot )$ represents the process of training the network with the weights $P^{\alpha+1} = \{P^{\alpha}_1,P^{\alpha+1}_3\}$. The performance metric of the network  with the weights $P^{\beta} = \{P^{\beta}_1,P^{\beta}_3\}$   on validation data is the fitness of the individuals $(k_1^j,k_2^j,...,k_n^j)$. During the training process of Equation (\ref{equation4}), the weights $P^{\alpha+1} = \{P^{\alpha}_1,P^{\alpha+1}_3\}$ of the network are updated by the learning rate $\eta_0 -\eta_1 < 0$, and the weights $P^{\alpha+1} = \{P^{\alpha}_1,P^{\alpha+1}_3\}$ of the network are updated by the learning rate ${{\eta }_{1}}$. The learning rate is constrained as:

\begin{equation}
    \eta_0 -\eta_1 < 0
    \label{equation5}
\end{equation}

The fitness calculation process can be expressed as:

\begin{equation}
    \begin{aligned}
        & {{P}^{\alpha +1}}=f\left( {{P}^{\alpha }} \right) \\ 
       & {{P}^{\beta }}\text{=}g(P_{1}^{\alpha },P_{3}^{\alpha +1},{{\eta }_{0}},{{\eta }_{1}}) \\ 
       & \text{subject to } {{\eta }_{0}}-{{\eta }_{1}} < 0 
    \end{aligned}
    \label{equation6}
\end{equation}

In order to analyze the population distribution in the NE-SGD approach, we define $\nabla$ to represent the difference between the maximum and minimum of the individual fitness in the NE-SGD population. In addition, $\nabla$ also represents the range of individuals in the population. When $\nabla$ is larger, the search range of the algorithm is wider and the algorithm can also achieve satisfying performance. In Subsection \uppercase\expandafter{\romannumeral4}-D, the distribution of NE-SGD population is analyzed in detail. 

\subsection{A hierarchical clustering-based suppression approach}

In the process of the evolutionary operations in NE-SGD, a local suppression algorithm is developed to improve the population diversity. The details of the algorithm are shown in Algorithm \ref{algorithm2} adjacent blocks in the network have similar feature extraction performance. Solution exploration among similar individuals consumes a large number of computer sources, and is easy to fall into local optimums. To improve the population diversity for high searching efficiency, similar individuals in the population need be partially suppressed during the evolution. In this way, the optimization ability of EN-SGD can be enhanced. The algorithm combines the cluster analysis with the niche suppression technology based on the crowding mechanism \cite{R38}. The clustering algorithm divides the population into multiple sub-populations and maintains its diversity at the same time. The niche method based on the crowding mechanism suppresses the individuals with poor fitness values of sub-populations. Those sub-populations are evolved by the clustering algorithm and eliminate some individuals in the population.

\begin{algorithm}[h]
    \caption{Hierarchical clustering distance suppression algorithm }
    \label{algorithm2}
    \begin{algorithmic}[1]    
    \REQUIRE Current population $K$ on parent individuals and springs individuals, the population $K$ size $2m$.
    \STATE $ O = \text{\o} $ ,$C = K$;
    \FOR{$i = 1,2,3,...,2m$}
        \FOR{$j = 1,2,3,...,2m$}
            \STATE $O(i,j) = d(k_i,k_j)$, $d(k_i,k_j)$ reflects the Hamming distance between individuals;
            \STATE $O(j,i)= O(i,j)$;
        \ENDFOR
    \ENDFOR
    \STATE Set the current cluster number $q=2m$;
    \WHILE{ $q<m$}
        \STATE Find the nearest two clusters $C_{i^*}$ and  $C_{j^*}$;
        \STATE $C_{i^*} = C_{i^*}+C_{j^*}$;
        \FOR{$j =j^*+1,j^*+2,..,q$}
            \STATE $C_j \leftarrow C_{j-1}$;
        \ENDFOR
        \STATE $O(j^{*},j^{*})=\text{\o}$;
        \FOR{$j=1,2,...,q-1$}
            \STATE $O(i^*,j) =d(C_{i^*},C_{j})$,$O(j,i^*)=O(i^*,j)$;
        \ENDFOR
        \STATE q=q-1;
    \ENDWHILE
    \STATE The individual fitness of each cluster $\left\{{{C}_{1}}, {{C}_{2}} , \ldots,{{C}_{k}}, \ldots ,{{C}_{q}} \right\}$ can be represented as $E_{C_k} = \{ E_{C_k^1},E_{C_k^2},...,E_{C_k^p}\}$. And ${{C}^{{}}}\text{=}\left\{ {{C}_{1}},{{C}_{2}},\ldots ,{{C}_{k}},\ldots ,{{C}_{q}} \right\}$,$k\text{=}0,1,2\ldots q$, the size of ${{C}_{k}}$ is represented as $p$;
    \FOR{$ i =1,2,...,q $}
        \IF{$|E_{C_i}|>3$}
            \STATE Reduce the lowest fitness individual $E^* = E_{C_i}^{\min} $ in each cluster, $E^* = E^* \times 10^{-2}$;
        \ENDIF
    \ENDFOR
    \ENSURE The new population fitness  $E^{*}_{C_k}=\{ E_{C_k^1},E_{C_k^2},...,E_{C_k^p}\}$.
    
    \end{algorithmic}
\end{algorithm}

In the evolutionary operations of EN-SGD, the individual is composed of binary codes. Therefore, the similarity between individuals is measured by Hamming distance. The binary codes of individuals are consisted of two states. We use $h_i=(x_{i1},x_{i2},...,x_{im})$ represent the $i\text{th}$ individual and ${{h}_{j}}=({{x}_{j1}},{{x}_{j2}},\ldots ,{{x}_{jm}})$ represent the $j\text{th}$ individual. Then, the distance between $h_i=(x_{i1},x_{i2},...,x_{im})$ and ${{h}_{j}}$ can be represented as Equation (\ref{equation7}).

\begin{equation}
    d(h_i,h_j)  =\sqrt{\sum\limits_{l=1}^{m}(x_{h_i^l}-x_{h_j^l})^2}
    \label{equation7}
\end{equation}

According to the distance between individuals based on Equation (\ref{equation7}), we can obtain the similarity between individuals. Essentially, the population composed of offspring individuals and parent individuals are clustered into $q$ clusters $\left\{ {{C}_{1}},{{C}_{2}},\ldots ,{{C}_{k}},\ldots ,{{C}_{q}} \right\}$ (see lines 2-20). The fitness of each individual in ${{C}_{k}}$ is represented as $E_{C_k} = \{ E_{C_k^1}, E_{C_k^2},...,E_{C_k^p}\}$, and $p$ is the size of cluster ${{C}_{k}}$. As $p>3$, the fitness of each individual in ${{C}_{k}}$ can be suppressed according to Equation (\ref{equation8}). $|\cdot|$ represents the size of ${{C}_{k}}$.

\begin{equation}
    \begin{aligned}
        & {{E}^{*}}=E_{{{C}_{k}}}^{\min }\times {{10}^{-2}} \text{if}|{{E}_{{{C}_{i}}}}|>3 \\ 
        & \text{subject to} E_{{{C}_{k}}}^{\min }=\min \{{{E}_{C_{k}^{1}}},{{E}_{C_{k}^{2}}},...,{{E}_{C_{k}^{p}}}\} \\        
    \end{aligned}
    \label{equation8}
\end{equation}

\section{Experiment Design}

To verify the effectiveness of the proposed EN-SGD approach, a series of experiments are designed and performed. Because the proposed NE-SGD framework aims that the loss function of a DNN converges to a satisfying minimum, so five experiments are performed in this paper: 1) investigating the performance metric of different NE-SGD optimized CNNs on different datasets, 2) compare and analyze the performance of the proposed framework on different datasets with existing CNNs, 3) analyzing the convergence of NE-SGD, 4) analyzing the distribution of individual fitness values in the population, and 5) inspecting the efficiency of each component of the proposed framework. In this section, the selected CNN models and benchmark datasets, as well as the parameter settings for these experiments are presented.

\subsection{Peer Competitors}

To demonstrate the effectiveness of the proposed algorithm, various peer competitors are selected for comparison. Three state-of-the-art CNNs with hand-designed architectures are selected and trained on four datasets. The chosen CNNs are ResNet \cite{R35}, DenseNet121 \cite{R36}, and DPN92 \cite{R37}. Two different ResNet versions are used: ResNet with the depth of 18 (ResNet18) and depth of 34 (ResNet34). All CNNs are trained under the cross-entropy criterion and batch normalized. To show the superiority of the NE-SGD framework, the experiments are designed to train CNNs through SGD or NE-SGD separately and compare the classification accuracies of the four models through SGD or NE-SGD.

\subsection{Benchmark datasets}

The CIFAR10, CIFAR100, SVHN, and Aliyun-10500 are selected as the benchmark datasets \cite{R32}, as these datasets are widely selected for evaluating the performance of developed DNNs.

CIFAR10 is a classification with a 10-category natural object, containing a training dataset of 50,000 images and a test dataset of 10,000 images. Each image in CIFAR10 has the dimension of $32 \times 32$ pixels. There is an equal number of samples in each category of the training dataset roughly, while each category has the exact same number of images in the test dataset.

\begin{figure*}
    \centering
    \subfigure[Image examples of three categories in CIFAR10]{
        \begin{minipage}[b]{0.48\textwidth}
        \includegraphics[width=1\textwidth,height =2.0cm]{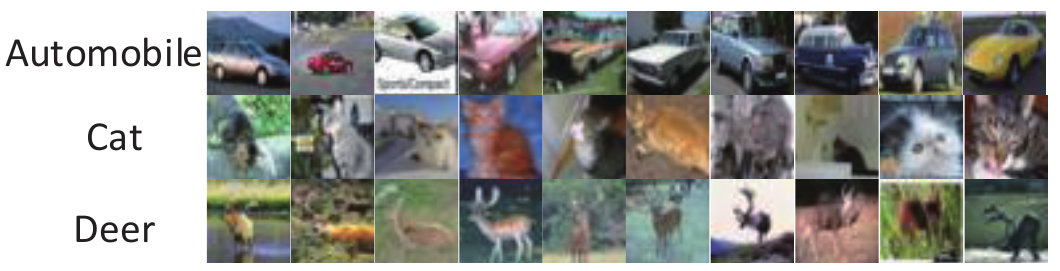}
        \end{minipage}
    }
    \subfigure[Image examples of three categories in CIFAR100]{
        \begin{minipage}[b]{0.48\textwidth}
        \includegraphics[width=1\textwidth,height =2.0cm]{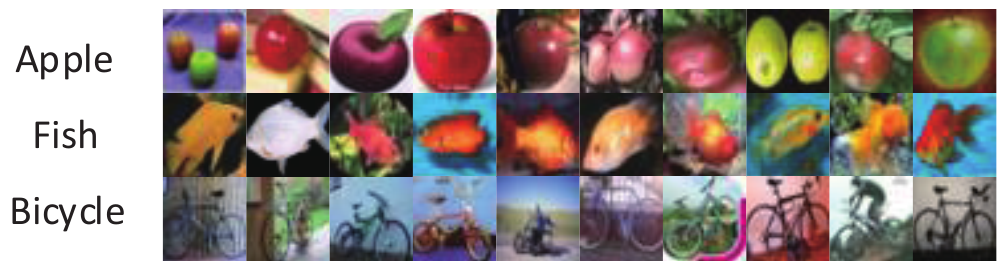}
        \end{minipage}
    }
    \subfigure[Image examples of three categories in SVHN]{
        \begin{minipage}[b]{0.48\textwidth}
        \includegraphics[width=1\textwidth,height =2.0cm]{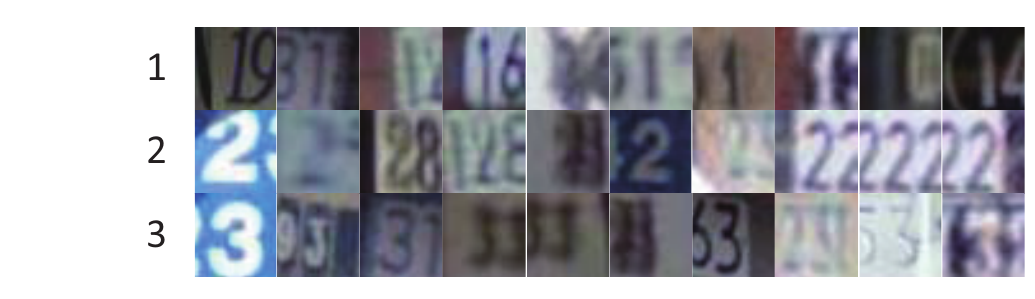}
        \end{minipage}
    }
    \subfigure[Image examples of three categories in Aliyun-10500]{
        \begin{minipage}[b]{0.48\textwidth}
        \includegraphics[width=1\textwidth,height =2.0cm]{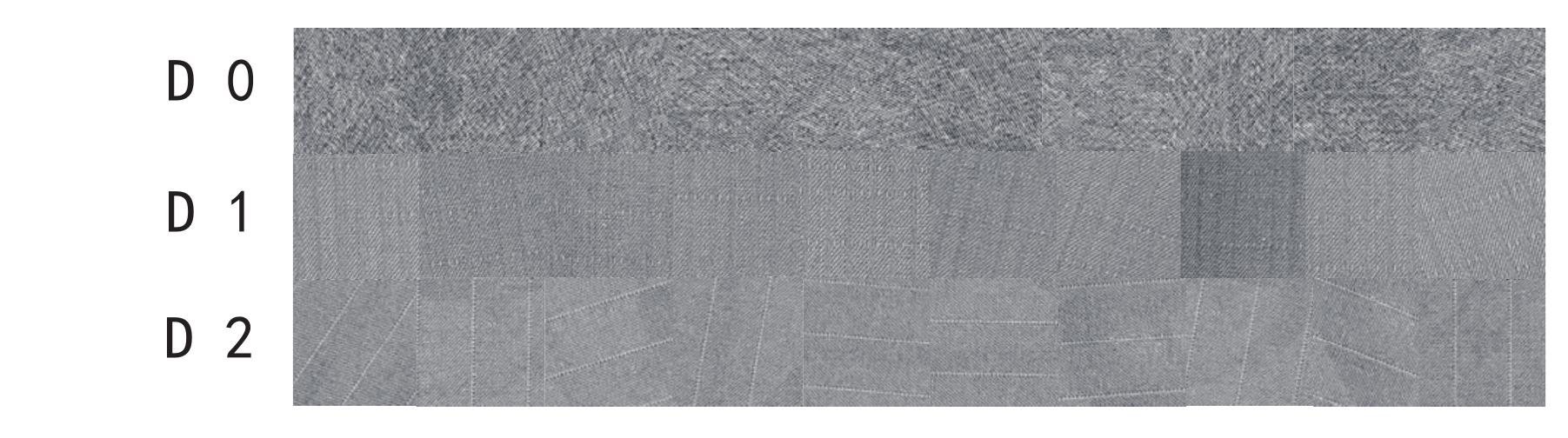}
        \end{minipage}
    }
    \caption{ Image samples in different datasets. Each row represents some samples from the same category with the category name in the left.} \label{fig:1}
\end{figure*}

CIFAR100 is simply like CIFAR10, except that it is 100-category. Because CIFAR100 contains the same number of images in the training dataset and test images as CIFAR10, each category in CIFAR100 has one-tenth of images as that in CIFAR10. CIFAR100 is much more challenging due to its larger number of classes for classification. Both CIFAR10 and CIFAR100 are chosen as the benchmark datasets. 

For every image of datasets of CIFAR10 and CIFAR100, the object to be classified ordinarily fills a few areas of the intact image. The size, area, and location of each object differ each other, even when they are from the equivalent category. The sample patterns of these datasets are shown in Fig. \ref{fig:1}, where each row expresses the objects from the same class and the label beginning in each row means the ground-truth of the similar object.

SVHN \cite{R33} is a large collection of $32 \times 32$ pixels RGB images. SVHN is a classification with a 10-category natural object, containing a training dataset of 73,257 images, a test dataset of 26,032 images and 53,131 extra training images. All of SVHN labeled digits are cropped from Street View images.

The Aliyun-10500 is collected from the public fabric classification competition (TianChi competition \cite{R44}) The dataset Aliyun-10500 consists of approximately 10,500 samples, including 9,000 defect samples and 2701,500 defect-free samples. The dataset contains seven categories of defects, each with 1,500 images. The types of defects include normal (defect-free), hole, crack, broken pick, broken end, stain, and felter.

\subsection{Parameter Settings}

Here, the parameter settings for training neural networks are given. Based on the general training setting of CNNs, the stochastic gradient descent method is often used to optimize networks. The batch size is set to 128; the weight decay is set as same as $5\times 10^{-4}$, each network is trained for 350 epochs. The learning rate is set as Algorithm \ref{algorithm1}. All the experiments are conducted on two GPUs with NVIDIA GeForce RTX 2080Ti and two CPUs with Intel Xeon Silver 4214. The codes of NE-SGD are written in Python 3.6.9 and Pytorch 1.3. The parameter settings are summarized in Table \ref{tab:test}. In order to further illustrate the influence of some randomness of SGD on NE-SGD, we designed a repeatability experiment based on SGD. In the experiment, the networks are trained 100 times by SGD under the same setting parameters as Table \ref{tab:test}.

\begin{table}
    \centering
     \caption{A summary of the parameter settings.}
     \begin{tabular}{cc}
      \toprule
      Parameter Settings & Parameter Value  \\
      \midrule
     batch size & 128  \\
     weight decay & $5\times 10^{-4}$  \\
     training epoch & $350$  \\
     generation number & $30$  \\
     population size & $5$  \\
     crossover probability & $0.9$  \\
     mutation probability & $0.1$  \\
      \bottomrule
     \end{tabular}
     \label{tab:test}
\end{table}

\section{Experiment results}

In the designed experiments, we first investigate the performance of the proposed NE-SGD approach implemented within different CNN models in terms of their classification accuracies. Next, the curve of accuracy-iteration to analyze the classification accuracy distribution of individual during the search process of the NE-SGD framework. Moreover, we also designed an ablation experiment for the effectiveness of Algorithm \ref{algorithm2}. In addition, we designed a comparative experiment to analyze the distribution of classification accuracy during 100 repetitive training of CNNs through SGD and further illustrate that the performance of NE-SGD is not related to the randomness of SGD.

\subsection{Performance of different NE-SGD-optimized CNNs on different datasets}

The performance of both NE-SGD implementations and only SGD implementations within the four CNNs is evaluated and compared on CIFAR10, CIFAR100, SVHN, and Aliyun-10500 datasets (Table \ref{Table2}). On each dataset, all CNNs trained with NE-SGD achieve better accuracies than those trained with only SGD. On CIFAR10, the accuracies increase by from $0.4\%$ (DPN92+NE-SGD) to $1.68\%$ (ResNet34+NE-SGD). It can be seen that the more complex the structure of a CNN is, the lower performance improvement the CNN can achieve. On CIFAR100, the ResNet34+NE-SGD achieves the highest performance improvement ($2.53\%$), while the DenseNet121+NE-SGD achieves the lowest performance improvement ($1.52\%$). Compared with the performance achieved on CIFAR10, the accuracies of all CNNs improve more on the more complex dataset (CIFAR100) by optimized with NE-SGD. On SVHN, ResNet18+NE-SGD achieves  a performance improvement of $0.99\%$. Particularly, on Aliyun-10500, ResNet18-based NE-SGD achieves $98.09\%$ classification accuracy, which is $3.54\%$ higher than that of the only SGD implementation based on ResNet18. Clearly the NE-SGD implementations outperform the only SGD implementations within all four representative CNN architectures on all datasets. The experiments on SVHN and Aliyun-10500 are to verify the universality of NE-SGD, so only one CNN model is selected for the evaluation. 

\begin{table*}
    \centering
    \caption{The classification accuracy (\%) of NE-SGD implementations with different CNN models on the four datasets.}
      \begin{tabular}{ccccc}
      \toprule
      DataSet & CNN model & Accuracy & CNN model (NE-SGD implementation) & \multicolumn{1}{l}{Accuracy} \\
      \midrule
      \multirow{4}[2]{*}{CIFAR10} & ResNet18 & 93.14 & ResNet18+NE-SGD & 94.82 \\
            & ResNet34 & 93.78 & ResNet34+NE-SGD & 95.05 \\
            & DenseNet121 & 95.04 & DenseNet121+NE-SGD & 95.61 \\
            & DPN92 & 95.16 & DPN92+NE-SGD & 95.56 \\
      \midrule
      \multirow{4}[2]{*}{CIFAR100} & ResNet18 & 75.27 & ResNet18+NE-SGD & 77.5 \\
            & ResNet34 & 75.78 & ResNet34+NE-SGD & 78.31 \\
            & DenseNet121 & 77.46 & DenseNet121+NE-SGD & 78.98 \\
            & DPN92 & 78.08 & DPN92+NE-SGD & 79.71 \\
      \midrule
      SVHN  & ResNet18 & 97.24 & ResNet18+NE-SGD & 98.23 \\
      \midrule
      Aliyun-10500 & ResNet18 & 94.61 & ResNet18+NE-SGD & 98.09 \\
      \bottomrule
      \end{tabular}%
    \label{Table2}%
  \end{table*}%

\begin{table*}
  \centering
  \caption{Comparisons between the proposed framework and the peer competitors in terms of the classification accuracy (\%).}
    \begin{tabular}{ccccc}
    \toprule
          & \multicolumn{1}{c}{CIFAR10} & \multicolumn{1}{c}{CIFAR100} & \multicolumn{1}{c}{SVHN} & \multicolumn{1}{c}{Aliyun-10500} \\
    \midrule
    ResNet(depth=20) \cite{R35} & \multicolumn{1}{c}{91.25} & \multicolumn{1}{c}{-} & \multicolumn{1}{c}{-} & \multicolumn{1}{c}{-} \\
    ResNet(depth=32) \cite{R35} & \multicolumn{1}{c}{92.49} & \multicolumn{1}{c}{-} & \multicolumn{1}{c}{-} & \multicolumn{1}{c}{-} \\
    DenseNet(depth=100) \cite{R36} & \multicolumn{1}{c}{94.23} & \multicolumn{1}{c}{76.21} & \multicolumn{1}{c}{98.33} & \multicolumn{1}{c}{-} \\
    Network in Network \cite{R39} & \multicolumn{1}{c}{91.19} & \multicolumn{1}{c}{64.32} & \multicolumn{1}{c}{97.65} & \multicolumn{1}{c}{-} \\
    FractalNet \cite{R40} & \multicolumn{1}{c}{94.78} & \multicolumn{1}{c}{77.7} & \multicolumn{1}{c}{97.99} & \multicolumn{1}{c}{-} \\
    VIN-Net \cite{R13} & \multicolumn{1}{c}{90.93} & \multicolumn{1}{c}{-} & \multicolumn{1}{c}{-} & \multicolumn{1}{c}{92.57} \\
    ESGD \cite{R29}  & \multicolumn{1}{c}{92.48} & \multicolumn{1}{c}{-} & \multicolumn{1}{c}{-} & \multicolumn{1}{c}{-} \\
    \midrule
    ResNet18+NE-SGD & \multicolumn{1}{c}{\textbf{94.82}} & \multicolumn{1}{c}{\textbf{77.75}} & \multicolumn{1}{c}{98.23} & \multicolumn{1}{c}{\textbf{98.09}} \\
    \bottomrule
    \end{tabular}%
  \label{Table3}%
\end{table*}%

  Model performance on CIFAR10 and CIFAR100 reflects a limitation of NE-SGD. NE-SGD outperforms SGD when the network and the dataset are in low complexity. When the structure of optimized CNN and data are complex, the performance of NE-SGD is limited. NE-SGD can achieve a good performance when DNN architecture is in low complexity.

\subsection{Performance comparison with existing CNN models on different datasets}

As shown in Table \ref{Table3}, ResNet18 with NE-SGD outperforms all the peer competitors designed for CIFAR10 and CIFAR100. Specifically, ResNet18 with NE-SGD achieves the classification accuracy of approximately $0.04\%$ higher than FractalNet, and even $3.57\%$ higher than ResNet (depth=20) on CIFAR10. At the same time, ResNet18 with NE-SGD achieves the performance metric of $1.8\%$ higher than ESGD on CIFAR10.  On CIFAR100, ResNet18 with NE-SGD shows significantly higher classification accuracy than DenseNet (depth=100), Network in Network, and FractalNet. On SVHN, ResNet18 with NE-SGD shows slightly lower classification accuracy than DenseNet (depth=100), while higher classification accuracy than Network in Network and FractalNet. In addition, ResNet18 with NE-SGD achieves the classification accuracy of approximately $5.52\%$ higher than VIN-Net on Aliyun-10500. Compared with peer competitors, ResNet18 network architecture is simpler. NE-SGD only optimizes the loss function of ResNet18 to obtain a classification accuracy comparable to that of its competitors.

\subsection{Convergence analysis of NE-SGD}

\begin{figure*}
    \centering
    \subfigure[SGD-optimized ResNet34 on CIFAR10]{
        \begin{minipage}[b]{0.48\textwidth}
        \includegraphics[width=0.98\textwidth]{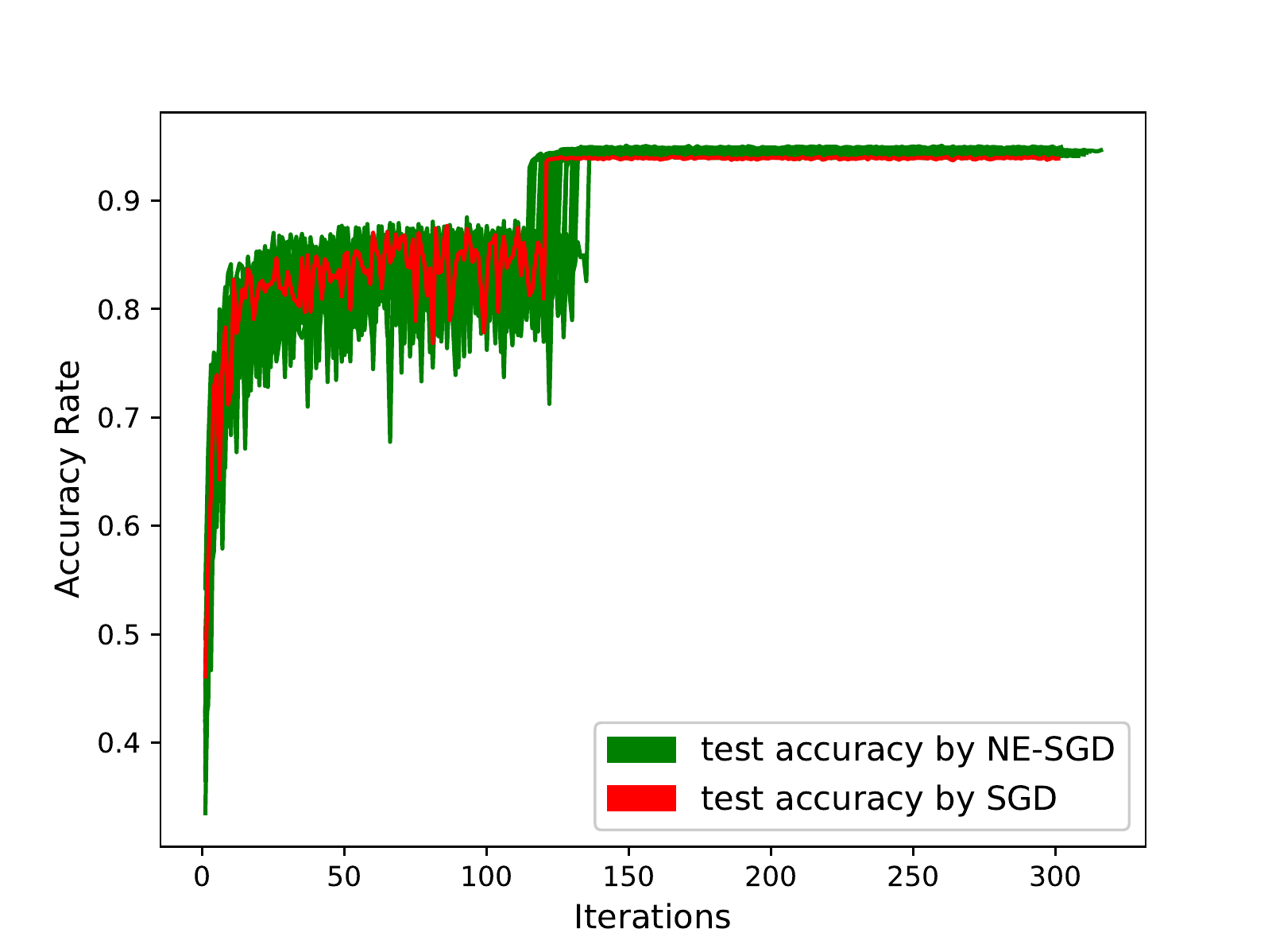}
        \end{minipage}
    }
    \subfigure[SGD-optimized ResNet34 on CIFAR100]{
        \begin{minipage}[b]{0.48\textwidth}
        \includegraphics[width=0.98\textwidth]{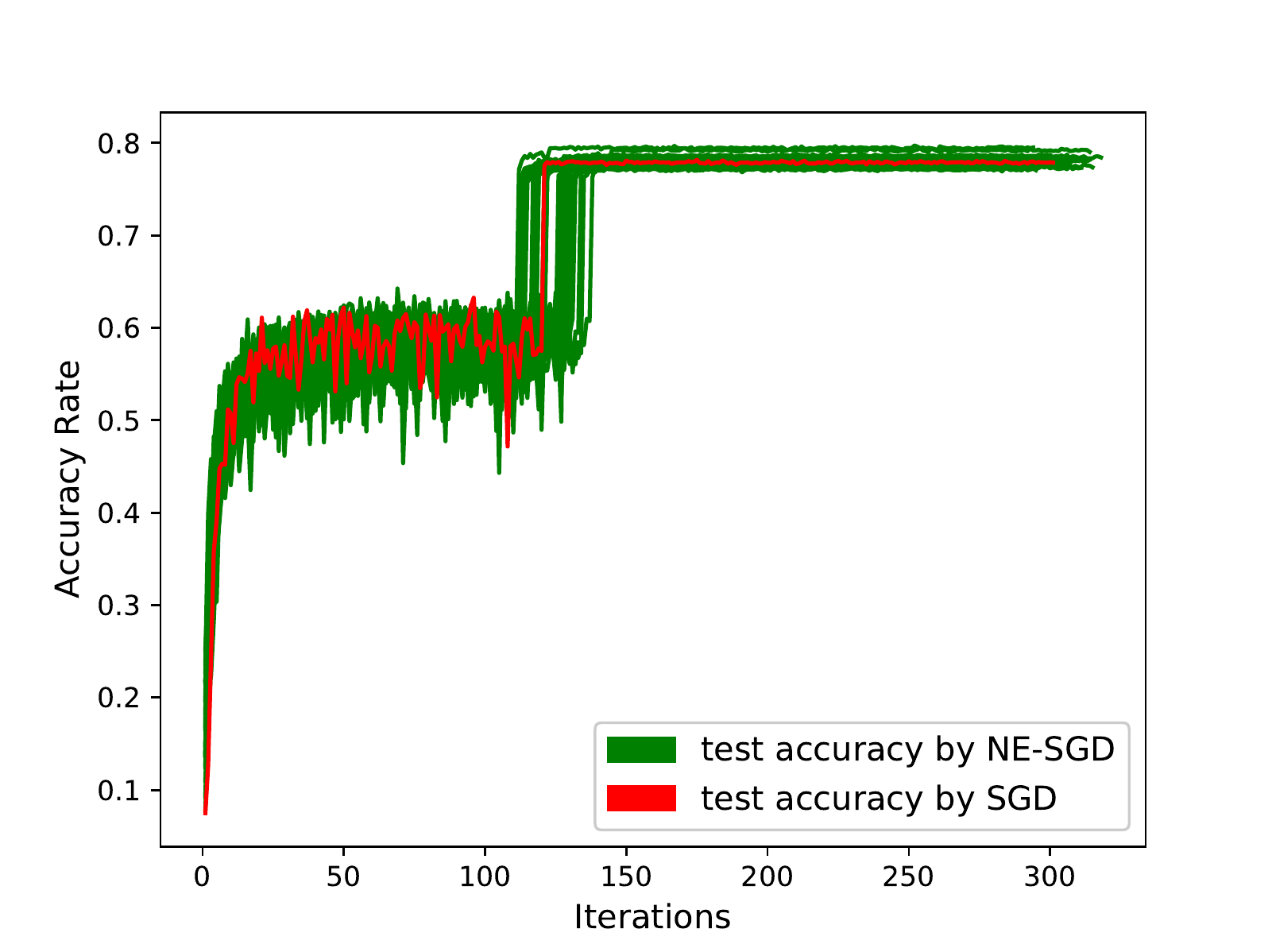}
        \end{minipage}
    }
    \caption{Convergence comparison between NE-SGD and SGD on CIFAR10 and CIFAR100, implemented within ResNet34.} \label{fig:2}
\end{figure*}

Fig. \ref{fig:2} shows the accuracy trend with NE-SGD on the dataset of CIFAR10 and CIFAR100. The red curve represents the trend of test accuracy in the process of training neural networks by SGD. The green curve represents the trend of test accuracy for all individuals in NE-SGD and shows the test accuracy change during training for all individuals in the NE-SGD population. These green curves are distributed on both sides of the red curve, which indicates that NE-SGD has some randomness in the search process. In some cases, it is invalid for NE-SGD to replace parts of neural network weights. The above can also be explained by the fact that the loss function of each individual in NE-SGD does not converge to a minimum. At present, the search of NE-SGD is a black-box process, which relies on non-gradient neuroevolution to complete the convergence of the loss function in the neural network. Neuroevolution uses test accuracy as a fitness value to ensure that the optimization process follows the direction of optimal test accuracy. In the process of NE-SGD searching for the satisfactory minimum, the increase of network complexity will bring many challenges to the optimization process. Therefore, we develop Algorithm 2 to increase the diversity of the population. The NE-SGD approach enhanced by Algorithm 2 obtains better performance.

\subsection{Population fitness analysis of NE-SGD}

Table \ref{table_3} reflects the distribution of NE-SGD population fitness in different CNNs. This table helps to reveal how NE-SGD works in different networks. $E(\nabla_{1}\nabla_{2})$ is defined as the average of $\nabla$ (see Subsection \uppercase\expandafter{\romannumeral2}-C for the definition) on ResNet18 and ResNet34. In addition, $E(\nabla_{3},\nabla_{4})$ represents the average of $\nabla$ on DenseNet121 and DPN92. Based on the results on CIFAR10 in Table 4, $E(\nabla_{1}^{C10},\nabla_{2}^{C10})=1.3\%$ and $E(\nabla_{3}^{C10},\nabla_{4}^{C10})=0.8\%$. The complexity of ResNet18 and ResNet34 is less than DenseNet121 and DPN92. It is inferred from the above that the search space of NE-SGD is wider in the network with lower complexity. Also, on CIFAR100, $E(\nabla_{1}^{C100},\nabla_{2}^{C100})=2.25\%$, and $E(\nabla_{3}^{C100},\nabla_{4}^{C100})=2.2\%$. However, the data distribution complexity of CIFAR10 is higher than that of CIFAR100, and the search range of NE-SGD on CIFAR10 is greater than the search range on CIFAR100. Therefore, based on the above analysis, we can infer that NE-SGD is more suitable for situations with a simple network structure and complex data distribution.

\begin{table*}
    \centering
    \caption{The comparison of box chart values about population statistics by the accuracy (\%) with ResNet18+NE-SGD, ResNet34+NE-SGD, DenseNet121+NE-SGD, and DPN92+NE-SGD based on CIFAR10 and CIFAR100.}
      \begin{tabular}{cccccc}
      \toprule
            &       & \multicolumn{1}{l}{ResNet18+NE-SGD} & \multicolumn{1}{l}{ResNet34+NE-SGD} & \multicolumn{1}{l}{DenseNet121+NE-SGD} & \multicolumn{1}{l}{DPN92+NE-SGD} \\
      \midrule
      \multirow{5}[2]{*}{CIFAR10} & minimum & 93.4  & 93.8  & 94.9  & 94.6 \\
            & lower quartile & 93.9  & 94.4  & 95.1  & 95 \\
            & median & 94.2  & 94.6  & 95.3  & 95.1 \\
            & upper quartile & 94.5  & 94.8  & 95.4  & 95.3 \\
            & maximum & 94.8  & 95    & 95.6  & 95.5 \\
      \midrule
      \multirow{5}[2]{*}{CIFAR100} & minimum & 75.6  & 75.8  & 76.9  & 77.3 \\
            & lower quartile & 76.4  & 76.6  & 77.7  & 78.2 \\
            & median & 76.7  & 77.1  & 78.1  & 78.3 \\
            & upper quartile & 76.9  & 77.2  & 78.4  & 78.6 \\
            & maximum & 77.6  & 78.3  & 78.9  & 79.7 \\
      \bottomrule
      \end{tabular}%
    \label{table_3}%
  \end{table*}%


  \begin{table}
    \centering
    \caption{The classification accuracy (\%) with and without the suppression algorithm.}
      \begin{tabular}{cccc}
          \toprule
      Network & \multicolumn{1}{c}{DataSet Name} & \multicolumn{1}{c}{With Suppression} & \multicolumn{1}{c}{Without Suppression} \\
      \midrule
      \multicolumn{1}{c}{\multirow{2}[1]{*}{ResNet18}} & \multicolumn{1}{c}{Cifar10\newline{}+cutout} & 95.60  & 95.50  \\
            & Aliyun-10500 & 98.09  & 97.90  \\
      \bottomrule
      \end{tabular}%
    \label{table_4}%
  \end{table}%

  \subsection{Ablation experiment results}

  Table \ref{table_4} shows the effect of the hierarchical cluster-based suppression algorithm in NE-SGD. Ablation experiments of the suppression algorithm are performed through ResNet18 on CIFAR10 and Aliyun-10500. To better show the performance improvement of the algorithm, we use cutout \cite{R42} on CIFAR10 to regularize ResNet18, improve the performance of NE-SGD, and further analyze the effect of the algorithm. On CIFAR10, it improves performance by $0.1\%$. On Aliyun-10500, it improves performance by $0.19\%$. The purpose of the algorithm is to improve the overly similar weight updating among individuals and creatively mine individuals who are beneficial to the improvement of network performance in the new weight update combinations. The ability to update weights for over-similar situations in the algorithm cannot be measured by indicators of population distribution, so the effect of the algorithm can only be evaluated by the improvement of algorithm performance. Based on the above analysis, the algorithm improve the overly similar weight updating among individuals in EN-SGD.

  Table \ref{table_5} shows the classification accuracies achieved by the networks trained 100 times by SGD. In the experiment, some random factors brought by SGD do not significantly affect the test accuracies of these models, and the test accuracies are maintained in a relatively stable interval. As the network ResNet18 is trained 100 times through SGD, the classification accuracy fluctuates by $0.72\%$, while the classification accuracy of the network ResNet34 fluctuates by $0.92\%$. The classification accuracy of the network ResNet18 based on NE-SGD is improved by $1.68\%$, and the classification accuracy of the network ResNet34 is improved by $1.72\%$. It can be seen from the above experiment that NE-SGD still improves network performance, although affected by SGD randomness. The results in Table \ref{table_5} further validate the effectiveness of NE-SGD.

  \begin{table}
    \centering
    \caption{The comparative experiments of the classification accuracy (\%) for NE-SGD. All the networks are trained by 350 epochs with SGD. Each model is trained 100 times after initialization of the same parameters.}
      \begin{tabular}{ccc}
      \toprule
            & ResNet18 & ResNet34 \\
      \midrule
      minimum & 93.13 & 93.52 \\
      lower quartile & 93.36 & 93.84 \\
      median & 93.42 & 93.95 \\
      upper quartile & 93.51 & 94.05 \\
      maximum & 93.85 & 94.46 \\
      standard deviation & 0.15  & 0.17 \\
      \bottomrule
      \end{tabular}%
    \label{table_5}%
  \end{table}%

\section{Conclusion}

In this paper, we develop a framework that combines neuroevolution with stochastic gradient descent for optimizing the loss function of DNNs. The goal of the proposed framework is to converge the loss function to a satisfactory minimum. In the framework of NE-SGD, the weights of the DNNs are encoded into a fixed-length binary string. Based on the binary string and neuroevolution, the network is retrained and the loss function of the network is converged to the satisfactory minimum. In order to address the overly similar weight updating among individuals and improve the diversity of the population, a hierarchical cluster-based suppression algorithm is proposed. In the experiments, the effectiveness of NE-SGD through Resnet18, ResNet34, DenseNet121, and DPN92 is tested on CIFAR10, CIFAR100, SVHN, and Aliyun-10500. Compared with the existing approaches, NE-SGD further improves the performance of DNNs.

This work highlights the combined method based on neuroevolution and SGD that further optimizes the loss function of DNN. Although combined with the effective SGD, the process of neuroevolution implies more consumption of computational resources, which leads to long processing time. NE-SGD is more useful in situations where DNNs performance is highly demanded. Our future work aims at reducing the processing time of NE-SGD, for example, through the deployment in high-performance computational platforms (e.g., \cite{R43}). In addition, our NE-SGD is only validated on the datasets of image classification. Future work will extend the proposed framework in other fields, such as object detection and instance segmentation.

\section*{Acknowledgments}

This work was supported in part by the National Natural Science Foundation of China (nos. 61806051, 61903078), and Natural Science Foundation of Shanghai (20ZR1400400, 19ZR1402300).

\ifCLASSOPTIONcaptionsoff
  \newpage
\fi



%

\bibliographystyle{IEEEtran}
\bibliography{cas-refs}

\end{document}